\documentclass{article}

\usepackage[numbers]{natbib}
\usepackage[preprint]{neurips_2022}

\usepackage[utf8]{inputenc} 
\usepackage[T1]{fontenc}    
\usepackage{hyperref}       
\usepackage{url}            
\usepackage{booktabs}       
\usepackage{amsfonts}       
\usepackage{nicefrac}       
\usepackage{microtype}      

\usepackage{amsmath}
\usepackage[dvipsnames]{xcolor}

\usepackage[utf8]{inputenc}

\usepackage{graphicx}
\usepackage{subcaption}
\usepackage{algorithm2e}
\usepackage{multirow}
\usepackage{wrapfig}

\title{Goal-Conditioned Supervised Learning with Sub-Goal Prediction}

\author{%
  Tom Jurgenson \\
  Faculty of Electrical Engineering \\
  Technion\\
  \texttt{tomj@campus.technion.ac.il} \\
  examples of more authors
  \And
  Aviv Tamar \\
  Faculty of Electrical Engineering \\
  Technion\\
  \texttt{avivt@technion.ac.il} \\
}

\begin{document}

\maketitle

\begin{abstract}
  Recently, a simple yet effective algorithm -- goal-conditioned supervised-learning (GCSL) -- was proposed to tackle goal-conditioned reinforcement-learning. 
  GCSL is based on the principle of hindsight learning: by observing states visited in previously executed trajectories and treating them as attained goals, GCSL learns the corresponding actions via supervised learning.
  However, GCSL only learns a goal-conditioned policy, discarding other information in the process.
  Our insight is that the same hindsight principle can be used to learn to predict goal-conditioned \textit{sub-goals} from the same trajectory. Based on this idea, we propose Trajectory Iterative Learner (TraIL), an extension of GCSL that further exploits the information in a trajectory, and uses it for learning to predict both actions and sub-goals. 
  We investigate the settings in which TraIL can make better use of the data, and discover that for several popular problem settings, replacing real goals in GCSL with predicted TraIL sub-goals allows the agent to reach a greater set of goal states using the exact same data as GCSL, thereby improving its overall performance.
\end{abstract}

\section{Introduction}



Ushered by advances in deep-learning in recent years, reinforcement learning (RL) has become a successful framework for solving sequential decision making problems~\cite{mnih2013playing, silver2016mastering, silver2018general, ibarz2021train}.
Goal-conditioned RL (GC-RL) is a variant of the general RL problem where an agent is tasked with fulfilling a goal that is given to it as input, instead of maximizing rewards~\cite{schaul2015universal, andrychowicz2017hindsight}. 
In principle, GC-RL can be solved as a regular RL problem using general purpose RL algorithms, with a reward function that is high only when the goal is met.
However, due to sparsity of this reward structure, RL is difficult to apply successfully~\cite{nair2018visual, charlesworth2020plangan}.
Crafting a suitable dense reward for such tasks, on the other hand, requires significant domain knowledge.
    
Recently, a simple algorithm for GC-RL, goal-conditioned supervised-learning (GCSL), was proposed~\cite{Ghosh2019LearningTR}. 
GCSL works by inferring in hindsight about which goals would have been fulfilled by a trajectory, regardless if it succeeded in reaching the intended goal or not. 
Moreover, unlike other popular hindsight methods such as hindsight experience replay (HER, \cite{andrychowicz2017hindsight}), GCSL optimizes a supervised-learning objective instead of an RL objective, making GCSL easier to train in practice, as common RL objectives, which are based on policy gradients or temporal difference learning, are known to be very sensitive to hyperparameters~\cite{henderson2018deep}. 


GCSL, however, comes with its own limitations. As we found, the sampling of targets in GCSL is biased towards learning short sub-trajectories of the data, making it difficult for GCSL to correctly predict actions for goals that require a long trajectory to reach.
Furthermore, learning to predict actions often disregards important ``geometric'' state-space information, such as the proximity between two states, that is very useful in GC-RL.
To tackle these issues we propose Trajectory Iterative Learner (TraIL), an extension of GCSL, which learns to predict a complete trajectory from the current state towards the goal.
We use TraIL's trajectory prediction for replacing the goal input of GSCL with a sub-goal, selected from the predicted trajectory to the goal. This effectively shortens the horizon for the GCSL action predictor, allowing it to predict actions with higher accuracy, so long as the sub-goal prediction is accurate enough.
Key in our approach is the observation that we can learn to predict the trajectory \textit{using the exact same data} that is available to GCSL. This allows us to build on the already established GCSL machinery when devising our algorithm.

In our empirical investigation, we seek to understand when TraIL can outperform GCSL. We handcraft environments to study this, and also evaluate performance on popular benchmarks. We find that in most cases TraIL leads to better goal coverage than GCSL, without requiring additional data.



We briefly summarize our contributions:
\begin{itemize}
    \item We show that GCSL is biased towards learning actions for closer goals.
    
    \item We propose TraIL, a sub-goal based extension of GCSL, and explore 
    regularization techniques and architecture design choices for implementing TraIL successfully.
    
    \item An empirical study of the benefits and limitations of our approach, as well an ablation study of the different TraIL components.

\end{itemize}

\section{Related work}

The sub field of GC-RL dates back to the classical goal-reaching RL algorithm by Kaelbling~\cite{kaelbling1993learning}.
More recent ideas, such as universal value function approximation (UVFA~\cite{schaul2015universal}), incorporated deep-learning in GC-RL effectively. 
To address the issue of rewards sparsity, the idea of \textbf{hindsight} was developed to re-evaluate past failed experience as successes. 
Some hindsight methods, such as HER~\cite{andrychowicz2017hindsight}, rely on the algorithm being off-policy and inject the hindsight transitions into the replay buffer, while others apply importance-sampling corrections within the RL algorithm~\cite{rauber2017hindsight}.

Several previous studies suggested using sub-goals to solve GC-RL tasks~\cite{pertsch2020long, jurgenson2020sub, parascandolo2020divide, chane2021goal}.
A key difference between these approaches and our work is that we, similarly to GCSL, only require a supervised learning objective, while all previous sub-goal based approaches optimized an RL objective, which can be more difficult to train. 

Recently, several works~\cite{janner2021offline, chen2021decision} used attention in the form of Transformers~\cite{vaswani2017attention} in order to learn future rewards with a supervised learning objective, without using sub-goals. 
We believe that these methods could also be improved to use sub-goals, based on ideas similar to the ones presented here.




\section{Problem Formulation and Background}

Let $M$ be a goal-conditioned MDP $(S,G, A, P, T, \rho_0)$, where $S$ and $A$ are the state and action spaces (either discrete or continuous), with $G \subseteq S$ being the goal space. $P(s'|s, g, a)$
is the transition function, where we assume that $P(g|g,g,\boldsymbol{\cdot})=1$ (i.e., goals are absorbing states), $\rho_0$ is the distribution over initial states and goals, and $T$ is the horizon of the episode.
The objective is to learn a (possibly stochastic) goal-conditioned policy $\pi(a | s, g)$ that successfully reaches goals in $T$ steps as follows:
\begin{equation}
    J_{goal-reaching}(\pi) = \mathbf{E}_{s_0, g \sim \rho_0, s_t\sim P_{\pi}(s| s_{t-1})}{\Big[\mathbf{I(s_T == g)} \Big]},
\end{equation}
where $P_{\pi}(s | s_{t-1}) = \sum_{a\in A}{P(s| s_{t-1}, a)\pi(a| s_{t-1})}$ and $\mathbf{I}$ is the indicator function.

\textbf{GCSL:} 
The main idea in GCSL\cite{Ghosh2019LearningTR} is that, in-hindsight, states visited by the agent could be considered as goals, even if they were not the intended goals when the agent executed its policy.
This self-supervised policy optimization is carried out via a maximum likelihood objective on the actions given the achieved goals.
Similarly to other online RL algorithms, GCSL interleaves policy optimization steps (described below) with collecting new data and appending it to a \textit{replay buffer} $D$. However, unlike other prevalent RL methods, the learning objective in GCSL is purely supervised (no value learning or estimation).
Formally, let $\tau = [s_0, a_0, \dots s_{T-1}, a_{T-1}, s_T]$ be a trajectory from the buffer where $\tau_s(i)$ and $\tau_a(i)$ denote the $i$'th state and action in $\tau$.
GCSL samples $i \sim U(1, T-1)$ which determines the current state $\tau_s(i)$ and action $\tau_a(i)$. 
The goal is then selected from the preceding visited states $j \sim U(i+1, T)$. The optimization objective in GCSL is:
\begin{equation}\label{eq:j-gcsl}
    J_{gcsl}=\mathbf{E}_{\tau\sim D, i \sim U(1, T-1), j \sim U(i+1, T)}{\Big[\log \pi(\tau_a(i) | \tau_s(i), \tau_s(j)) \Big]}.
\end{equation}

\section{GCSL is biased towards learning short trajectories}\label{sec:gcsl_is_biased}

Our main observation, and the motivation for our work, is that GCSL's learning objective has an inherent bias toward goals that are \textit{short horizoned}, i.e., a short trajectory is required to reach them. Thus, we expect that for goals that require a longer trajectory to reach, the predictions made by GCSL will be less accurate. 
In this section, we provide a mathematical justification this observation.

The GCSL objective (Eq.~\ref{eq:j-gcsl}) can be written as an optimization on trajectory-suffixes. Denote by $\tau^i$ the suffix of $\tau$ that starts in state $\tau_s(i)$. The objective can be written as:
\begin{equation*}
    J_{gcsl}=\mathbf{E}_{\tau\sim D, i \sim U(1, T-1), j \sim U(1, T - i)}{\Big[\log \pi(\tau^i_a(1) | \tau^i_s(1), \tau^i_s(j)) \Big]}
\end{equation*}
We ask, what is the probability that GCSL updates a target that is exactly $j=K$ steps away, and we denote this event as $U(K)$.
Define $p_k$ as the probability of selecting a trajectory-suffix that is of length $k$. Thus, we can write $U(K)$ as the sum:
\begin{equation}\label{eq:gcsl-u-k}
    \mathbf{Pr}\big(U(K)\big) = \sum_{k \ge K}{\frac{p_k}{k}} = \sum_{k \ge K} {\frac{p_k}{k}} + \sum_{k < K} {p_k \cdot 0}.
\end{equation}

The first equality follows since $\tau^i$ is required to be of length at least $K$ (the restriction in the summation), and then specifically selecting a goal that is exactly $K$ steps away (GCSL selects $j$ under a uniform distribution).
The leftmost expression just adds zero elements for $k<K$.

For targets that are exactly $K+1$ steps away, plugging in Eq.~\ref{eq:gcsl-u-k},
\begin{equation*}
    \mathbf{Pr}\big(U(K+1)\big) = \sum_{k \ge K+1} {\frac{p_k}{k}} + \sum_{k < K+1} {p_k \cdot 0}.
\end{equation*}
Therefore, we have that $\mathbf{Pr}\big(U(K)\big) = \frac{p_K}{K}  + \mathbf{Pr}\big(U(K+1)\big)$, and since $p_K \ge 0$, we have that $\mathbf{Pr}\big(U(k)\big)$ is monotonically decreasing $\mathbf{Pr}\big(U(K)\big)\ge \mathbf{Pr}\big(U(K+1)\big)$.
This result is independent on the exact value of $p_k$, which can depend both on the dynamics $P$ and the data collection policy.

\begin{figure*}
\centering
\includegraphics[width=1.0\textwidth]{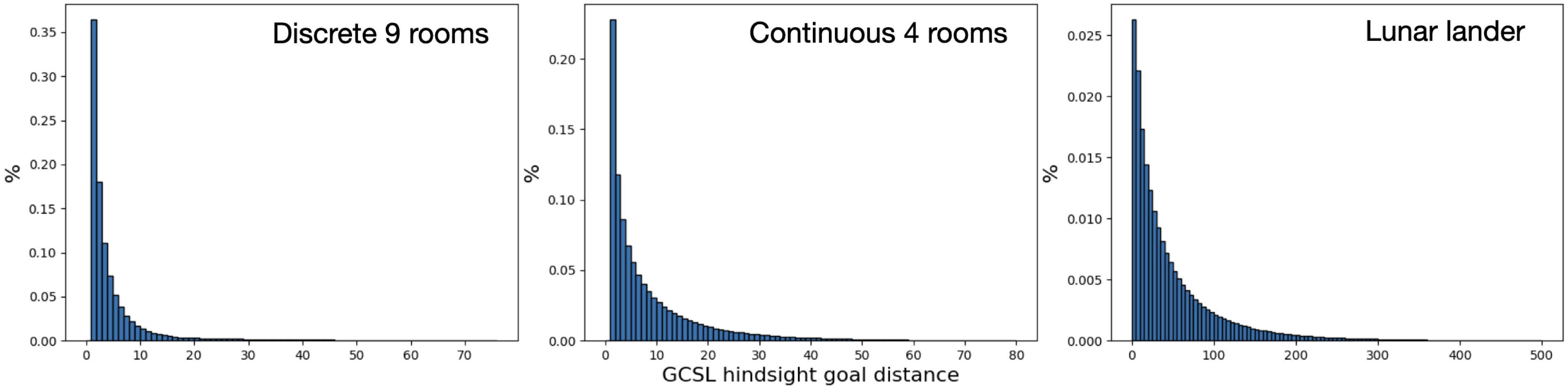}
\caption{Histograms of distances between start and goal states when computing the GCSL targets in the \textit{discrete 9 rooms} (left), \textit{continuous 4 rooms} (middle), and \textit{lunar lander} (right) environments. 
The right end of the X axis is the horizon $T$. 
Y axis is the percent of the bin from the total updates.
We see that GCSL implicitly learns targets for shorter sub-trajectories more frequently.
}
\label{fig:sub_trajectory_histogram}
\end{figure*}

We conclude that closer targets get updated more frequently in GCSL, biasing the model to learn more from data that corresponds to short horizoned goals.
We further illustrate this issue in Figure~\ref{fig:sub_trajectory_histogram}, where we plot a histogram of the lengths of sub-trajectories that GCSL learns from.
We see that most updates correspond to relatively short sub-trajectories suffixes, despite the fact that collected trajectories may, and often do, reach the maximal length $T$ (the right end of the X axis).

\section{The Trajectory Iterative Learner (TraIL) Algorithm}\label{sec:trail_algorithm}


In the previous section we discussed why GCSL is biased towards short-horizon trajectories.
Motivated by this observation, we aim to utilize short-horizon predictions to solve long-horizon ones. Our idea is that in some scenarios, learning future states leading to a goal, could be an easier task than directly learning the action to take at the first step of the trajectory. Once we have such a \textit{sub-goal} at hand, we can use it instead of the original goal when querying GCSL, potentially receiving a more accurate action prediction.

To realize this idea, during learning a GCSL model $\pi$, we concurrently learn a trajectory encoder $\pi_S(m | s,g,t)$ to predict the state between a current $s\in S$ and goal $g\in G$ states from previous data.
$\pi_S$ is indexed by $t\in[0,1]$ such that if $t=0$ the required result should be $m=s$, and if $t=1$ then it should be $m=g$. 
Furthermore, $\pi_S$ should generate subsequent states that are feasible according to $P$ (including the transition to $g$). 
If the above requirements hold for any two reachable $s$ and $g$, then we call the trajectory encoder \textit{consistent}.

Next, we describe how to train our algorithm, TraIL, in Section~\ref{subsec:trail-train}, and how to use  TraIL sub-goals to predict GCSL actions in Section~\ref{subsec:action-prediction}.

\subsection{TraIL optimization and training}\label{subsec:trail-train}
To train $\pi_S(m | s,g,t)$ we assume the same algorithmic settings as in GCSL, of an iterative process that shifts between collecting data and adding it to a replay buffer, and optimizing a supervised-learning objective.
The objective is based on log-likelihood to fit the data that was collected by GCSL so far during training (see below), simialr to Eq.~\ref{eq:j-gcsl}.
As the GCSL policy $\pi$ trains, TraIL $\pi_S$ trains in the background on the same data, 
i.e., we compute more update steps but consume the same data.
Our loss function 
for $\pi_S$ is given by:
\begin{equation}
    J_{sub-goal}=\mathbf{E}_{\tau\sim D, i \sim U(1, T-1), j \sim U(i+1, T), k \sim U(i, j)}{\Big[\log \pi_S(\tau_s(k) | \tau_s(i), \tau_s(j), \frac{k -i}{j-i}) \Big]}.
\end{equation}
The objective $J_{sub-goal}$ maximizes the likelihood of the visited state $\tau_s(k)$ according to its relative position in the sub-trajectory $t=\frac{k -i}{j-i}$.

Next, we explain our neural network architecture design. Additionally, for a state space where the Euclidean distance is meaningful, we propose a novel regularization technique to train TraIL.\footnote{Examples of state spaces where the Euclidean distance is not meaningful include states of categorical nature (like in logical planning problems), images, or domains with a discontinuity in the state space (e.g., an angle that jumps from $2\pi$ to $0$).}


\textbf{Architecture:}
Unlike the classical supervised learning settings where a reasonable assumption is a single data collection policy, in GCSL, where the policy iterates between optimization and data collection, this assumption no longer holds.
At every point in time the replay buffer contains data of several modalities corresponding to past versions of the current policy.
We thus use a model capable of representing several modalities -- Mixture Density Network (MDN~\cite{bishop1994mixture}), a neural network predicting a mixture of Gaussians, for $\pi_S$.
Let $K$ be the number of Gaussians, and $d$ be the state dimension. The network predicts $p \in \Delta_K$, $c_1 \dots c_K \in R^d$, and $\hat{\sigma_1} \dots \hat{\sigma_K}\in R^d$.
The output $p$ is the logits for selecting each mode, and the Gaussian $N_k$ for mode $k$ is $N(\mu_k = s+c_k, \sigma_k^2 = \exp{(2\hat{\sigma_k})})$ (we center the means around $s$).

For ease of notation, in the next sections we define:
Let $k^*(s,g,t)$ be the most likely mode ($\arg\max_{k\in [K]} (p)$) and $\mu_k(s,g,t)$ the mean of mode $k$ (emphasizing the input dependencies).
Also, let $\overline{k(s,g)}$ denote the mode that starts closest to $s$ and terminates closest to $g$ (for explicit description see the BestMode Algorithm~\ref{alg:best-mode} in the supplementary section).



\textbf{Regularization: } motivated by the properties of a \textit{consistent} trajectory encoder, we propose two regularization terms $J_{edge}$ and $J_{self-consistency}$ (see below), that bias $\pi_S$ towards smooth consistent trajectories. Both losses only require sampling states $s,g$ from the data-set $D$.
Our complete loss $J_{TraIL}$ for the trajectory encoder is (with $\alpha_{edge}$ and $\alpha_{self-consistency}$ are hyper-parameters):
\begin{equation}
    J_{TraIL} = J_{sub-goal} + \alpha_{edge} \cdot J_{edge} + \alpha_{self-consistency} \cdot J_{self-consistency}.
\end{equation}

\paragraph{Edge loss:} For a consistent encoder we require that $\pi_S(s,g,0) = s$ and $\pi_S(s,g,1) = g$ (with a slight abuse of notation as $\pi_S$ is stochastic). 
    Thus, we add the following loss that predicts the \textbf{location} of the start and goal:
    \begin{equation}\label{eq:trail-edge-loss}
        J_{edge}=\mathbf{E}_{s,g \sim D}{\Big[\| \mu_{k^*}(s,g,0) - s\|^2 + \| \mu_{k^*}(s,g,1) - g\|^2 \Big]}.
    \end{equation}
    
\begin{figure*}[h]
\centering
\includegraphics[width=0.5\textwidth]{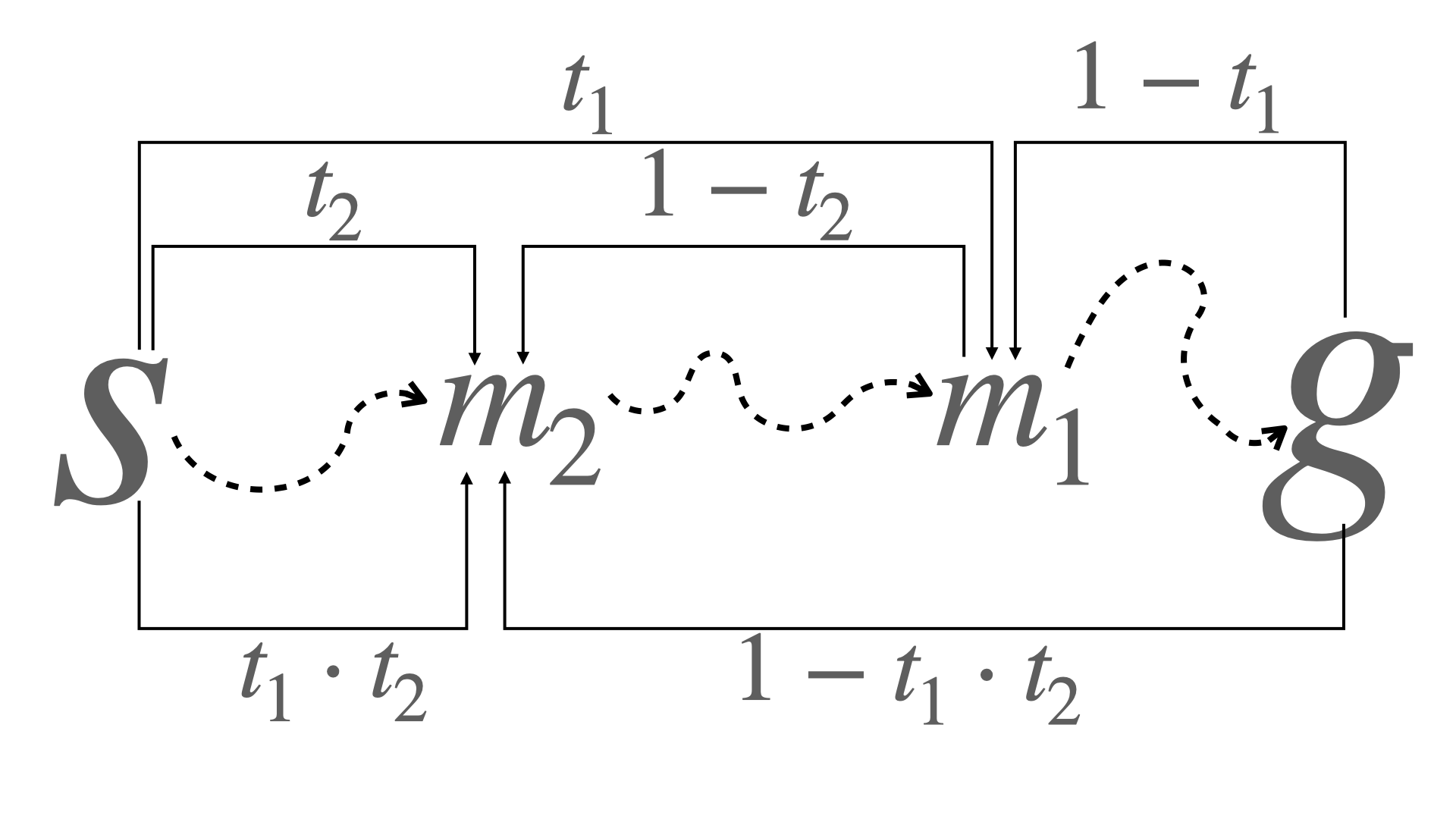}
\caption{
An illustration of the self-consistency regularization. 
We sample $m_1$ from $s$ and $g$, and $m_2$ from $s$ and $m_1$, and minimize the residual of predicting $m_2$ using $s$ and $g$ with an appropriate $t$ value.}
\label{fig:traj_encoder_reg}
\end{figure*}

\paragraph{Self-consistency loss:} For a consistent encoder, let $s_m$ be a state in trajectory $\tau$, all sub-trajectories of $\tau$ that contain $s_m$ should predict it (with appropriate $t$ values). 
    Namely, the predictions of $\pi_S$ for a trajectory $\tau$ and a sub-trajectory $\tau^i$ should agree on overlapping states (see Figure~\ref{fig:traj_encoder_reg} for a graphical illustration). 
    The ``self-consistency'' loss based on this motivation is:
    \begin{align}\label{eq:trail-self-consistency-loss}
        J_{self-consistency}&=\mathbf{E}_{s,g \sim D, t_1,t_2 \sim U(0,1)}{\Big[
        \|\mu_{k'}(s,g,t_1\cdot t_2) - m_2\|^2
        \Big]}\nonumber\\
        &=\mathbf{E}_{s,g \sim D, t_1,t_2 \sim U(0,1)}{\Big[
        \|\mu_{k'}(s,g,t_1\cdot t_2) - \mu_{k'}(s,\mu_{k'}(s,g,t_1),t_2)\|^2
        \Big]}
    \end{align}
    With $k'=k^*(s,g,t_1)$ being the mode that we regularize for.

\textbf{Trajectory post-processing: }
In the GC-RL setting, we are interested in reaching MDP states, therefore subsequent identical states in the data provide no value towards reaching any goal. 
Thus, when a trajectory $\tau$ contains subsequent identical states $\tau_s(i)= \tau_s(i+1)$ we trim the repeating state $\tau_s(i+1)$ (and action $\tau_a(i)$). 
The process repeats until all subsequent states are different and only then the trimmed $\tau$ is pushed into the replay buffer $D$. 
We found this pre-processing step to greatly enhance results (Section~\ref{sub-sec:minigrid-post-process}), and we apply it to both GCSL and TraIL in the experimental section.

\subsection{Predicting an action by first predicting a sub-goal}\label{subsec:action-prediction}

It was previously shown that divide-and-conquer prediction approaches are effective at solving long horizon predictions\cite{pertsch2020long, jurgenson2020sub, parascandolo2020divide, chane2021goal}.
Following these sub-goal prediction methods, when predicting an action for data collection, we query $\pi_S$ with $t=0.5$ and generate a sub-goal $m$ and then replace $g$ with $m$ in GCSL action prediction $a\sim \pi(a | s,m)$.\footnote{When the current time step $i$ approaches $T$ we instead choose a more "aggressive" goal $t=\max(0.5, \frac{i+1}{T})$}
During test-time, the only difference is that the most likely action is taken (highest logit if $A$ is discrete, or the mean of the most likely mixture component $\mu_{k^*}(s,g)$ if $A$ is continuous). Formally:

\textbf{GetAction($s,g,i$)}

\textbf{Inputs:} current and goal states $s, g \in S$, current time index $i$
\textbf{Output:} action $a\in A$
\begin{enumerate}
    \item $t=\max(0.5, \frac{i+1}{T})$
    \item $k\leftarrow\overline{k(s,g)}$
    \item $m \leftarrow \mu_{k}(s,g,t)$ when deterministic, or $m \sim N_{k}(s,g,t)$ if stochastic
    \item $a\sim \pi(s, m)$
\end{enumerate}


    
    
    
    
    
    

\section{Is action prediction via sub-goals always easier?}

\begin{figure*}
\centering
\includegraphics[width=0.45\textwidth]{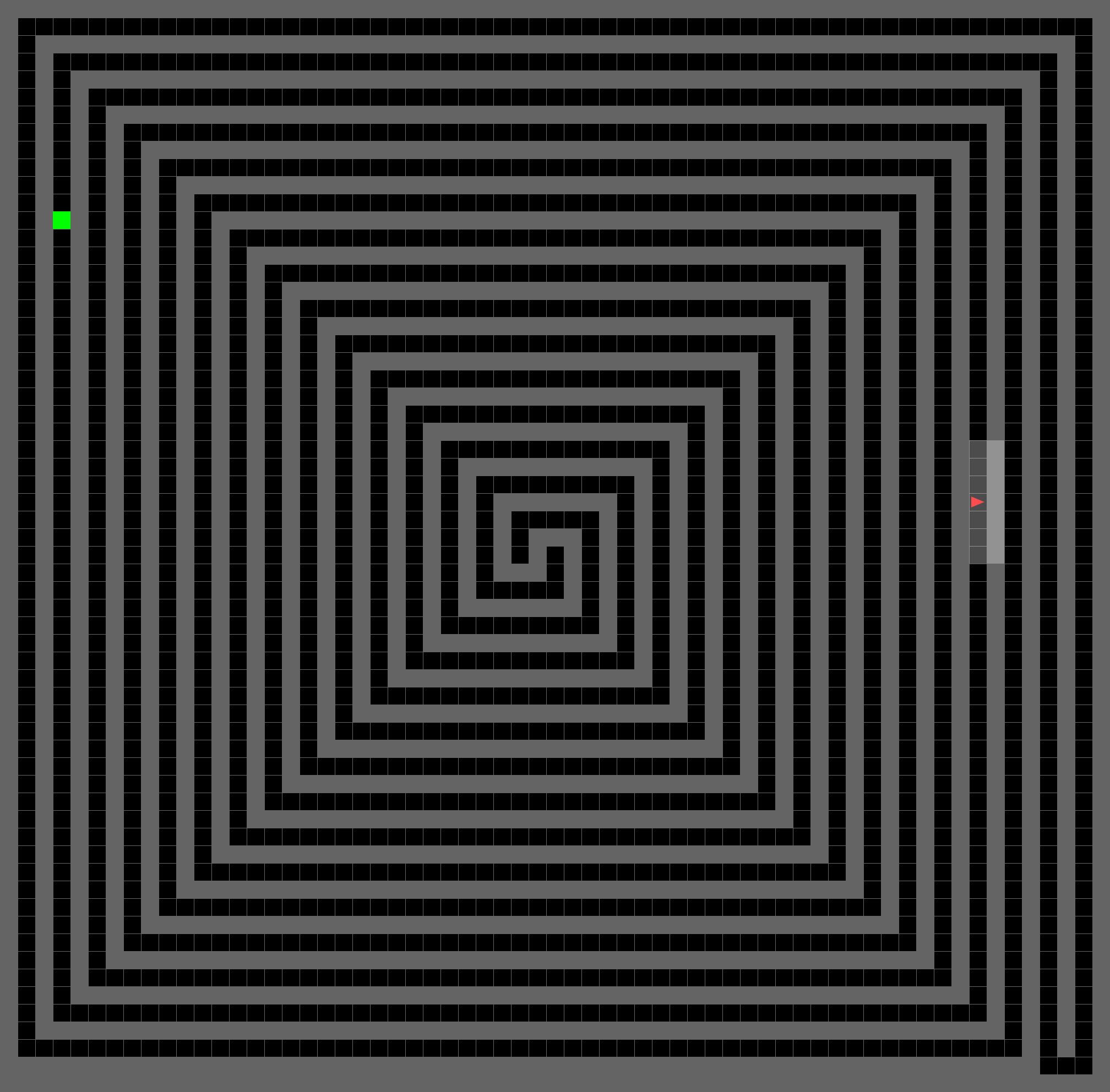}
\hfill
\includegraphics[width=0.45\textwidth]{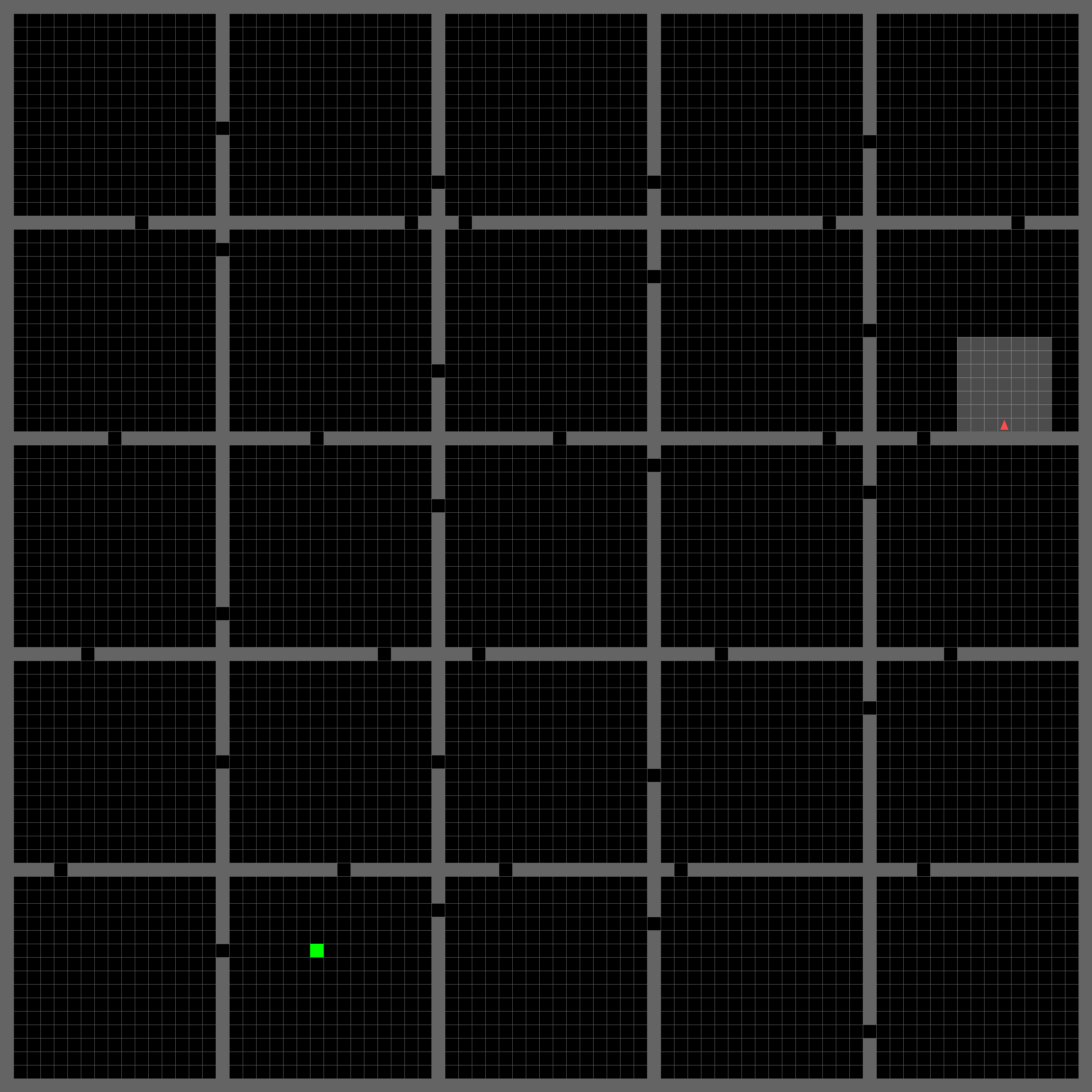}
\caption{The environments used in the behavioral-cloning experiments: \textit{double spiral} on the left and \textit{large rooms} on the right. 
The start (red triangle) and goal (green cell) change in each demonstration.
}
\label{fig:bc_envs}
\end{figure*}

The major difference between TraIL and GCSL is 
that GCSL learns actions, while TraIL learns sub-goal trajectories. In this section we investigate when learning sub-goals can be easier than predicting actions. In such cases, we expect TraIL to improve over GCSL.
Our main intuition, which also guided our choice of regularization terms earlier, 
is that TraIL should improve upon GCSL in MDPs where small errors in sub-goal predictions do not cause large changes in the predicted action.

To investigate this question, we design an experiment that side steps the data collection process common in RL scenarios, and instead we focus on behavioral cloning (BC~\cite{pomerleau1988alvinn}) scenarios where we are given a set of demonstrations, generated as the shortest path trajectories between random states and random goals, and must learn to imitate the demonstrations. 
This formulation allows us to investigate the prediction problem in isolation, without additional RL difficulties such as exploration.
We designed two specialized mini-grid environments for this investigation
(Figure~\ref{fig:bc_envs}).
The first scenario, \textit{large-rooms}, is a 25 rooms grid, with each room being 15x15 cells. Every room is connected to adjacent rooms by a single cell door. 
We expect sub-goals to be easy to predict in this domain, 
as prediction errors for sub-goals could be proportional to the room size without affecting the resulting action prediction.
The second scenario, \textit{double-spiral}, has 2 intertwined corridors that twist and change directions often. 
The corridors are a single cell wide, meaning that for every goal, exactly one action takes the agent in the correct direction, another one in the opposite direction, and the remaining two actions keep the agent in place. 
We hypothesize that learning sub-goals in this scenario would be more difficult, as two close states on different corridors correspond to very different trajectories towards the goal.


In this experiment, we compare action predictions between the GCSL model, i.e. $a \sim \pi(a | s,g)$, and the TraIL model that first predicts a sub-goal $m\sim \pi_S(m | s,g,t)$, and then predicts an action $a \sim \pi(a | s,m)$.
For TraIL, we compare three variants: $t=1.0,$ without regularization, and $t=0.5$ with and without regularization.\footnote{With regularization $\alpha_{edge}=\alpha_{self-consistency}=0.01$, without 0.} 
Success is measured by the accuracy of predicting the first action
of the shortest-path demonstrated trajectory,
for a fixed set of start-goals pairs that were not seen by the agents during training.
The results, in Table~\ref{tab:bc-experiments}, agree with our intuition.
In \textit{large rooms}, adding more features of TraIL improves the success rate of the model, while in \textit{double spiral} it worsens it.
\begin{table}[h]
\centering
\begin{tabular}{l|l|l|l|l}
              & GCSL & TraIL: $t=1$ & TraIL: $t=0.5$ & TraIL: $t=0.5$, reg \\ \hline
large rooms   & 0.741 $\pm$ [0.007]      & 0.742 $\pm$ [0.006]     & 0.758 $\pm$ [0.013]       & 0.771 $\pm$ [0.001]            \\
double spiral & \textbf{0.849 $\pm$ [0.010]}      & 0.795 $\pm$ [0.007]     & 0.637 $\pm$ [0.014]       & 0.626 $\pm$ [0.023]          
\end{tabular}
\caption{
Accuracy of GCSL vs. TraIL action predictions in the \textit{large-rooms} and \textit{double-spiral} domains. 
In TraIL columns, $t$ is the index of the sub-goal, and "reg" indicates that both edge and self-consistency losses were active in the optimization (see text for more details).
}
\label{tab:bc-experiments}
\end{table}

\section{Experiments}\label{sec:experimetns}

\begin{figure*}
\centering
\includegraphics[width=1.0\textwidth]{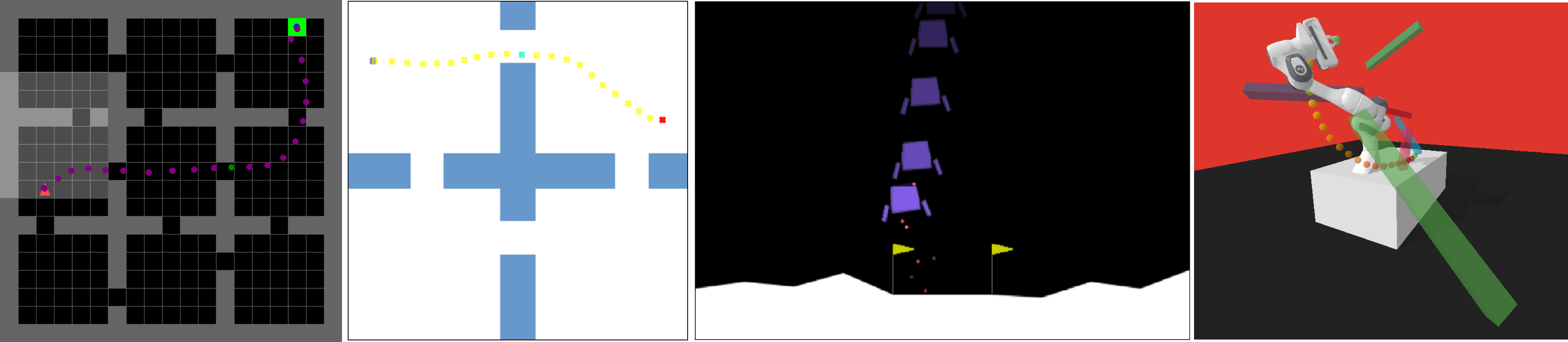}
\caption{GC-RL environments, and the executions of successful TraIL trajectories. 
From the left:  \textit{discrete 9 rooms}, \textit{continuous 4 rooms}, \textit{lunar lander}, and \textit{panda motion planning}.
}
\label{fig:gcrl_envs}
\end{figure*}


We now turn to study TraIL in the full GC-RL settings, and compare to GCSL.\footnote{See Section~\ref{appdx:technical-details} for full technical details.}
%
Our scenarios (Figure~\ref{fig:gcrl_envs}) include:  
    
    \textit{Discrete 9 rooms}: is a harder version of the minigrid 4 rooms problem, where 9 rooms are aligned in a 3x3 grid, and each room is 5x5 cells. 
    The doors to adjacent rooms are in random locations (not necessarily in the center of the wall).
    This scenario has both discrete state and action spaces, and the dynamics function is deterministic.
    We chose this relatively simple scenario as it is easy to visualize the resulting policies.
    
    \textit{Continuous 4 rooms}: a grid of 2x2 rooms similar to the previous scenario, except the state and action spaces are continuous and the dynamics function has 3 levels of difficulty \textit{no-noise}, \textit{moderate-noise}, and \textit{heavy-noise}, corresponding to noise levels of 0.0, 0.1, and 0.5.
    
    \textit{Lunar lander}: 
    Lunar lander is a scenario with challenging dynamics that also appeared in the original GCSL paper. \cite{Ghosh2019LearningTR} adapted lunar lander to the GC-RL setting as follows:
    (1) the goal is defined by positions (without velocities), and (2) the goal region is limited around the landing-pad flags.
    We further modify the settings to (1) use the velocities in goals and sub-goals,
    and (2) define three goal ranges: \textit{landing-pad} (as in \cite{Ghosh2019LearningTR}), \textit{close-proximity}, and \textit{wide-proximity}.
    The latter two goal regions 
    require the agent to perform a more sophisticated trajectory than just falling straight down.
    We denote the environment setting of \cite{Ghosh2019LearningTR} as \textit{Lunar-GCSL-settings}.
    
    \textit{Panda motion planning}: 
    We investigate a setting of neural motion planing (NMP), a challenging GC-RL task~\cite{qureshi2018motion, jurgenson2019harnessing}, in which a 7-DoF Franka Panda robot must navigate between 
    pole-like objects.\footnote{Usually, external demonstrations or other privileged information is required to allow RL agents to solve NMP with high success rates~\cite{jurgenson2019harnessing}.}


To evaluate our agent, we measure the success rates of both agents on a held-out set of start-goal queries.
Full results are shown in Tables~\ref{tab:experiments-main1} and~\ref{tab:experiments-main2}.

\begin{table}[h]
\centering
\begin{tabular}{c|c|ccc|c}
 & \multirow{2}{*}{\begin{tabular}[c]{@{}c@{}}discrete \\ 9 rooms\end{tabular}} & \multicolumn{3}{c|}{\begin{tabular}[c]{@{}c@{}}continuous\\ 4 rooms\end{tabular}} & \multirow{2}{*}{\begin{tabular}[c]{@{}c@{}}panda motion\\ planning\end{tabular}} \\
\multicolumn{1}{l|}{} &  & \multicolumn{1}{l}{no-noise} & \multicolumn{1}{l}{moderate-noise} & \multicolumn{1}{l|}{heavy-noise} &  \\ \hline
GCSL & \begin{tabular}[c]{@{}c@{}}0.567 $\pm$ {[}0.051{]}\end{tabular} & \begin{tabular}[c]{@{}c@{}}0.652 $\pm$ {[}0.015{]}\end{tabular} & \begin{tabular}[c]{@{}c@{}}0.616 $\pm$ {[}0.056{]}\end{tabular} & \begin{tabular}[c]{@{}c@{}}0.158 $\pm$ {[}0.003{]}\end{tabular} & 0.693 $\pm$ [0.017] \\
TraIL (ours) & \textbf{\begin{tabular}[c]{@{}c@{}}0.719 $\pm$ {[}0.059{]}\end{tabular}} & \textbf{\begin{tabular}[c]{@{}c@{}}0.913 $\pm$ {[}0.015{]}\end{tabular}} & \textbf{\begin{tabular}[c]{@{}c@{}}0.74 $\pm$ {[}0.039{]}\end{tabular}} & \begin{tabular}[c]{@{}c@{}}0.158 $\pm$ {[}0.005{]}\end{tabular} & \textbf{0.767 $\pm$ [0.006]}
\end{tabular}
\caption{
Success rates of GCSL and TraIL (ours) in \textit{discrete 9 rooms}, \textit{continuous 4 rooms}, and \textit{panda motion planning}. See main text for breakdown of results.
}
\label{tab:experiments-main1}
\end{table}

\begin{table}[h]
\centering
\begin{tabular}{c|c|ccc}
 & \begin{tabular}[c]{@{}c@{}}GCSL-settings\\ (no-velocities)\end{tabular} & \multicolumn{3}{c}{Extended state-space} \\
\multicolumn{1}{l|}{} & \multicolumn{1}{l|}{landing-pad} & \multicolumn{1}{l}{landing-pad} & \multicolumn{1}{l}{close-proximity} & \multicolumn{1}{l}{wide-proximity} \\ \hline
GCSL & \textbf{\begin{tabular}[c]{@{}c@{}}0.847 $\pm$ {[}0.019{]}\end{tabular}} & \begin{tabular}[c]{@{}c@{}}0.472 $\pm$ {[}0.149{]}\end{tabular} & \begin{tabular}[c]{@{}c@{}}0.479 $\pm$ {[}0.060{]}\end{tabular} & \begin{tabular}[c]{@{}c@{}}0.341 $\pm$ {[}0.054{]}\end{tabular} \\
TraIL (ours) & \begin{tabular}[c]{@{}c@{}}0.815 $\pm$ {[}0.018{]}\end{tabular} & \begin{tabular}[c]{@{}c@{}}0.585 $\pm$ {[}0.142{]}\end{tabular} & \textbf{\begin{tabular}[c]{@{}c@{}}0.706 $\pm$ {[}0.059{]}\end{tabular}} & \textbf{\begin{tabular}[c]{@{}c@{}}0.543 $\pm$ {[}0.013{]}\end{tabular}}
\end{tabular}
\caption{
Success rates of GCSL and TraIL (ours) in all \textit{lunar lander}. See main text for breakdown of results.
}
\label{tab:experiments-main2}
\end{table}


We start with the most crucial question - does TraIL improve upon GCSL? According to the results in Tables~\ref{tab:experiments-main1} and~\ref{tab:experiments-main2}, we observe that in most of our tested scenarios TraIL indeed offers significant success rate improvements compared to GCSL.\footnote{We provide additional experiments in the supplementary that demonstrate a similar conclusion.}
Even in the challenging high-dimensional \textit{panda motion planning} scenario, we find that TraIL is able to improve upon GCSL with a substantial gap.
A notable exception is \textit{lunar lander} in the GCSL settings, which we will discuss later on.
We next emphasize a few interesting findings. 




\paragraph{Effects of increasing noise in dynamics function:} in the \textit{continuous 4 rooms} we investigated the relationship between noise in the dynamics function and the algorithms' performance.
We observe that, as expected, both algorithms suffer due to increased noise, but TraIL still outperforms GCSL for the moderate noise level.
We hypothesize that since GCSL breaks under heavy noise, TraIL, which uses GCSL as a low-level policy, is also unable to perform well.
In general, we can expect that GCSL performing reasonably is a requirement for TraIL to succeed.

\paragraph{Exploration: } exploration is an important consideration in most GC-RL scenarios. 
GC-RL tasks are more challenging when the agent's initial state distribution does not cover the full support of state-goal pairs. 
In such cases, 
the agent must explore effectively, and incorporate new information 
into its learned model.

\textit{Lunar lander} is a good example for such a scenario. The agent starts above the flags and (in the GC-RL settings) needs to fly to specific goals.
Table~\ref{tab:experiments-main2} shows that TraIL manages to cover more goals than GCSL under the challenging \textit{close-proximity} and \textit{wide-proximity} goal regions.
We note, that because TraIL does \textbf{not} control the data collection policy, this gap must be due to better usage of the same collected data.

In Section~\ref{sec:cont-4-rooms-limited} in the supplementary material we experimented with allowing TraIL to control the data collection process with favorable results: 
we modified the \textit{continuous 4 rooms} scenario such that the agent can only start in the top left room. 
We saw that with GCSL data both algorithms had low success rates of only 0.25 for GCSL and 0.27 for TraIL. By using more data collected with TraIL, success rates increased to 0.45 for GCSL and 0.47 for TraIL (compared to a baseline that kept collecting data with GCSL and had drop in performance), suggesting that at least in some settings, using TraIL to collect data could lead to more performance gains. 
We leave a large scale investigation into this direction for future work.

\begin{figure*}
\centering
\includegraphics[width=1.0\textwidth]{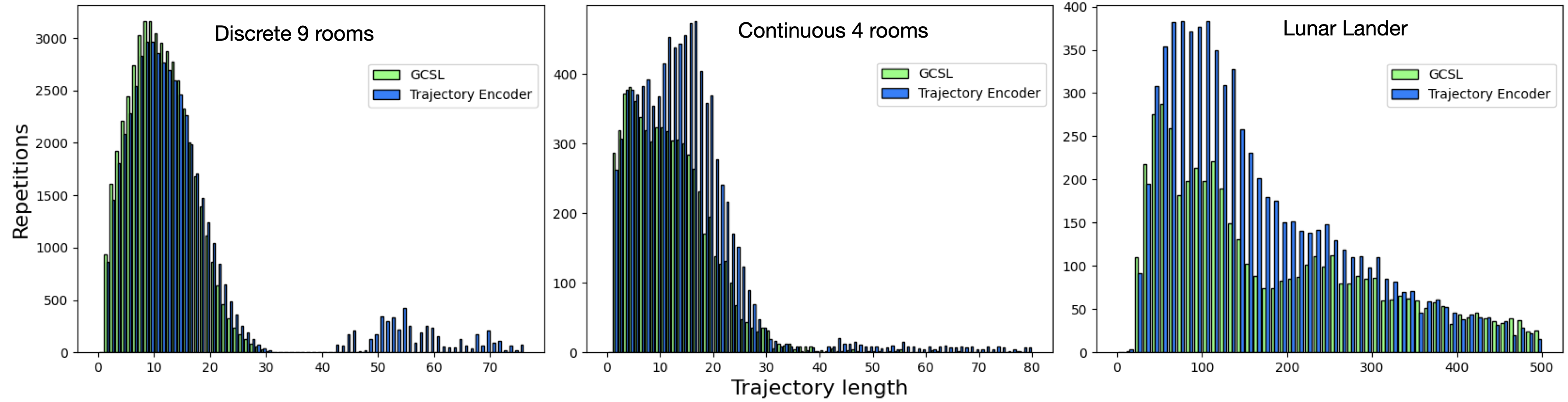}
\caption{Histogram of lengths of successful trajectories in the \textit{discrete 9 rooms} (left), \textit{continuous 4 rooms} (middle), and \textit{lunar lander} (right) environments.
X axis is the length of the successful trajectory, Y axis is the bin count for that length.
The histograms show that TraIL successes are more concentrated on long trajectories compared to GCSL.
}
\label{fig:success_by_length}
\end{figure*}

\paragraph{Length of successful trajectories:} in Section~\ref{sec:gcsl_is_biased} we hypothesized that GCSL is biased toward optimizing shorter sub-trajectories. 
We thus investigate the lengths of \textbf{successful} trajectories of both algorithms (Figure~\ref{fig:success_by_length}). 
Interestingly, the cases where TraIL succeeds and GCSL fails are not uniformly distributed, but concentrate on the longer trajectories. This observation confirms our intuition -- TraIL improves the performance of GCSL by allowing more accurate predictions for more distant goals.


\paragraph{Ablation study: } we used the \textit{discrete 9 rooms}, \textit{continuous 4 rooms}, and \textit{lunar lander} as a benchmark to test the key features in our method - the number of mixture components $K$, and the effects of regularization terms.
Our key finding are that adding a mixture usually helps, but a lower mixture count ($k=2$) seems like a good general purpose choice. 
For regularization, not adding regularization at all usually results in models of lower success rates, however, setting the regularization coefficients too high also negatively impacts performance. 
We saw that $\alpha_{edge}=\alpha_{self-consistency}=0.01$ is a good choice in the tested scenarios. See Section~\ref{sec:ablations} for more information.

\begin{figure*}
\centering
\includegraphics[width=0.33\textwidth]{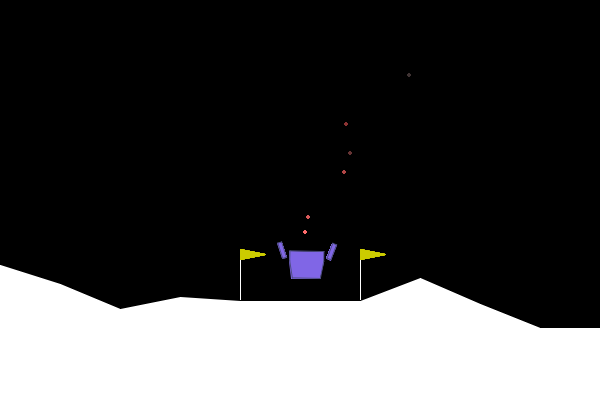}
\includegraphics[width=0.65\textwidth]{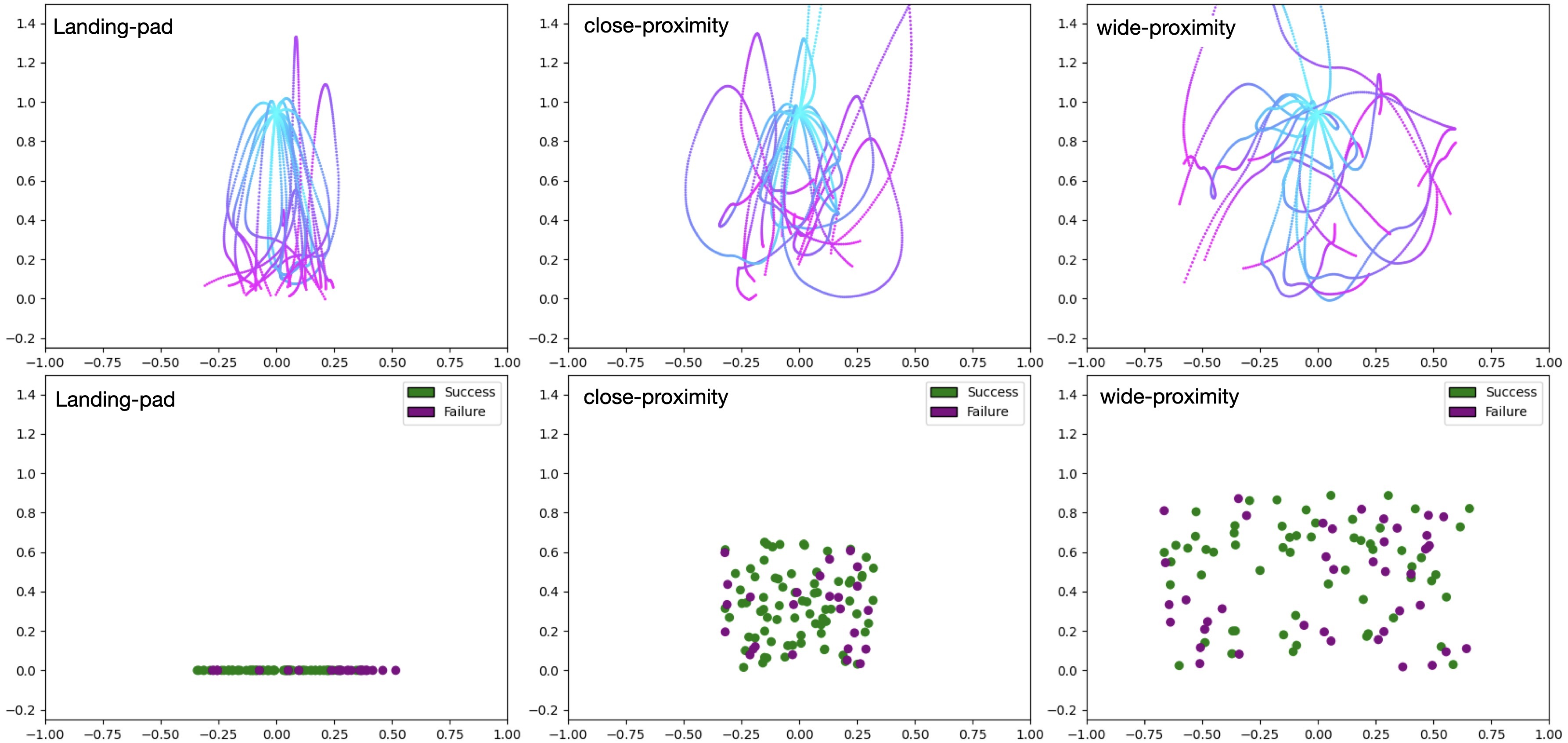}
\caption{
Lunar investigation: Left: GCSL agent in \textit{GCSL-settings} learns to fly head-first into the ground. 
Right top:
Goal visualization colored by success status of a TraIL agent, green for success and purple for failure. 
Right bottom: Visualization of full successful trajectories by TraIL, in each trajectory the agent starts located in the turquoise color, and as the trajectory evolves corresponding states transition to light pink.
Left: \textit{landing-pad}, middle: \textit{close-proximity}, and right \textit{wide-proximity}.
}
\label{fig:lunar_all}
\end{figure*}

\textbf{Lunar lander:} 
Upon close inspection of the \textit{Lunar-GCSL-settings} results, we saw that GCSL learns to fly head first into the ground. This undesirable behavior, which was reported in \cite{Ghosh2019LearningTR} as a success of reaching the goal, also makes reaching sub-goals harder, which explain the worse performance of  
TraIL in this experiment (see Figure~\ref{fig:lunar_all} left).

To solve this environment we made two modifications. 
First, we added velocities to sub-goals, which is crucial information for learning to reach a sub-goal in the correct heading for later continuing to reach the goal.
Second, we defined three goal regions: \textit{landing-pad}, \textit{close-proximity}, and \textit{wide-proximity} (see Figure~\ref{fig:lunar_all} top right) to diversify the data the agent sees during data collection.
Under these new settings, as shown in Table~\ref{tab:experiments-main2}, TraIL has superior performance compared to GCSL. In Figure~\ref{fig:lunar_all} top right we visualize the goal regions (green indicates success) and on the bottom right we plot successful trajectories (bottom) showing the diverse set of goals TraIL can reach. 
This experiment also highlights that TraIL requires a sub-goal representation with enough information for stable control (in this case - velocities).




\section{Discussion and Limitations}
In this work we presented TraIL, an extension of GCSL tackling the GC-RL problem. 
The key idea
in TraIL, is that GCSL's performance can be boosted by providing it with sub-goals, 
thereby reducing GCSL's prediction horizon.
TraIL learns sub-goals using the same data as GCSL itself.
We showed that for common scenarios, 
using TraIL leads to learning a better coverage of the state space. 
We also discussed TraIL's limitations -- 
it may not work well in domains where predicting a sub-goal is harder than predicting actions. Interestingly, in many benchmark tasks this was not the case.
An interesting future direction is to extend TraIL to visual domains, by incorporating ideas from visual planning.


\bibliography{example_paper}
\bibliographystyle{plain}


\newpage

\appendix

\section{Full results \textit{discrete 9 rooms}, \textit{lunar lander}}




In this section we report additional finding regarding the \textit{discrete 9 rooms} and the \textit{lunar lander} scenarios.
We found that in both environments when we diverge from the original GCSL training of updating the model once for every environment step, and instead train three times, the success rate significantly increases.
We therefore include the full results for both \textit{discrete 9 rooms} in Table~\ref{tab:minigrid-experiments} and for \textit{lunar lander} in Table~\ref{tab:lunar-experiments}. 
We can see that in both cases TraIL maintains a success rate gap compared to GCSL.

\begin{table}[h]
\centering
\begin{tabular}{l|ll|ll}
      & \multicolumn{2}{c|}{Post process=False}                                   & \multicolumn{2}{c}{Post process=True}                                   \\
      & \multicolumn{1}{c|}{updates=1}       & \multicolumn{1}{c|}{updates=3} & \multicolumn{1}{c|}{updates=1}       & \multicolumn{1}{c}{updates=3} \\ \hline
GCSL  & \multicolumn{1}{l|}{0.472 $\pm$ [0.012]}          & 0.553 $\pm$ [0.021]                         & \multicolumn{1}{l|}{0.567 $\pm$ [0.051]}           & 0.931 $\pm$ [0.010]                        \\
TraIL (ours) & \multicolumn{1}{l|}{\textbf{0.592 $\pm$ [0.020]}} & \textbf{0.69 $\pm$ [0.031]}                & \multicolumn{1}{l|}{\textbf{0.719 $\pm$ [0.059]}} & \textbf{0.99 $\pm$ [0.002]}              
\end{tabular}
\caption{
Success rates in \textit{discrete 9 rooms}. 
We compare GCSL and TraIL with and without the trajectory post-processing from Section~\ref{subsec:trail-train}, and with 1 and 3 updates per environment steps.
}
\label{tab:minigrid-experiments}
\end{table}

\begin{table}[ht]
\centering
\begin{tabular}{cc|c|ccc|}
\multicolumn{2}{l|}{\multirow{2}{*}{}} & \begin{tabular}[c]{@{}c@{}}GCSL-settings\\ (no-velocities)\end{tabular} & \multicolumn{3}{c|}{Extended state-space}           \\
\multicolumn{2}{l|}{}                  & landing-pad                                                             & landing-pad     & close-proximity & wide-proximity  \\ \hline
\multirow{2}{*}{updates=1}   & GCSL    & \textbf{0.847 $\pm$ [0.019]}                                                                   & 0.418 $\pm$ [0.247]           & 0.393 $\pm$ [0.060]          & 0.319 $\pm$ [0.047]          \\
                             & TraIL   & 0.815 $\pm$ [0.018]                                                                  & 0.450 $\pm$ [0.241] & \textbf{0.582 $\pm$ [0.089]} & \textbf{0.583 $\pm$ [0.057]} \\ \hline
\multirow{2}{*}{updates=3}   & GCSL    & -                                                                       & 0.472 $\pm$ [0.149]          & 0.479 $\pm$ [0.060]            & 0.341 $\pm$ [0.054]          \\
                             & TraIL   & -                                                                       & 0.585 $\pm$ [0.142] & \textbf{0.706 $\pm$ [0.059]} & \textbf{0.543 $\pm$ [0.013]}
\end{tabular}
\caption{
Success rates in \textit{lunar lander}. 
We compare GCSL and TraIL on several instances of the problem: \textit{GCSL-settings} the agent is only provided with positional 
}
\label{tab:lunar-experiments}
\end{table}

\section{Does post-processing help? A study in the \textit{discrete 9 rooms} scenario}\label{sub-sec:minigrid-post-process}

We investigate our scheme of post-processing episodes (Section~\ref{subsec:trail-train}), in the \textit{discrete 9 rooms} environment.
This environment is ideal as a test-bed for this feature since it is discrete and deterministic, thus repeating states are likely.
The results are summarized in table~\ref{tab:minigrid-experiments}, in the first row (labeled as "GCSL").
As table~\ref{tab:minigrid-experiments} demonstrates, post-processing of repeating states greatly improves performance. The time complexity of this operation is linear in the original episode, thus as cheap as as inserting the episode into the replay buffer, which amounts to negligible time costs. 
Therefore for the results in the main text (Section~\ref{sec:experimetns}) we present results for GCSL and TraIL with post processing applied (in effect also boosting the GCSL baseline).

\section{Ablation - mixture components and regularization term coefficients}\label{sec:ablations}

In this section we investigate how the GMM mixture count ($K$) and the edge and self-consistency coefficients ($\alpha_{edge}$ and $\alpha_{self-consistency}$) effect the success rates of TraIL.
To achieve that, we took a single trained GCSL model, fixed $\pi$ and only trained the TraIL policy $\pi_S$ according to the feature under investigation.
We trained on the \textit{discrete 9 rooms} with 1 update per environment step, \textit{cont 4 rooms}, and \textit{lunar} with \textit{close-proximity} goals, as these are variants of every scenario that have a noticeable gap to perfect success.
The success rate of the GCSL model is specified directly under the scenario name for reference. 

\subsection{Ablation I: effects of mixture components}

We start by investigating the effects of TraIL's mixture count $K$ on the success rates (Table~\ref{tab:mixture-ablation}). 
In both \textit{discrete 9 rooms} and in \textit{continuous 4 rooms} incorporating a mixture ($K>1$) is beneficial, and in the \textit{lunar lander} environment $K=1$ works marginally better.
We hypothesize that this due to the nature of the problem, in both room scenarios, there are many situations where a goal-reaching policy can take several directions to reach the goal, whereas in \textit{lunar lander} (with non-trivial dynamics), it is probably more difficult to find a diverse goal-reaching strategies to reach the same goal.

\begin{table}[ht]
\centering
\begin{tabular}{c|ccccc}
                                                                          & K=1                     & K=2                     & K=3                     & K=5                     & K=10                    \\ \hline
\begin{tabular}[c]{@{}c@{}}discrete 9 rooms\\ (GCSL: 0.6067)\end{tabular} & 0.704 $\pm$ [0.007] & 0.788 $\pm$ [0.013] & 0.757 $\pm$ [0.007] & 0.756 $\pm$ [0.032] & 0.738 $\pm$ [0.039] \\
\begin{tabular}[c]{@{}c@{}}cont 4 rooms\\ (GCSL: 0.6713)\end{tabular}   & 0.878 $\pm$ {[}0.009{]} & 0.923 $\pm$ {[}0.007{]} & 0.918 $\pm$ {[}0.009{]} & 0.921 $\pm$ {[}0.005{]} & 0.876 $\pm$ {[}0.041{]} \\
\begin{tabular}[c]{@{}c@{}}lunar lander\\ (GCSL: 0.5429)\end{tabular}  & 0.748 $\pm$ {[}0.005{]} & 0.745 $\pm$ {[}0.006{]} & 0.718 $\pm$ {[}0.014{]} & 0.705 $\pm$ {[}0.014{]} & 0.696 $\pm$ {[}0.008{]}
\end{tabular}
\caption{
Ablation of the mixture component $K$ in TraIL in various scenarios. 
GCSL success rate is listed under the scenario name for comparison.
}
\label{tab:mixture-ablation}
\end{table}

\subsection{Ablation II: self-consistency and edge loss}
Next we investigate the $J_{edge}$ and $J_{self-consistency}$ (Eq.~\ref{eq:trail-edge-loss} and \ref{eq:trail-self-consistency-loss}) by modifying their coefficients $\alpha_{edge}$ and $\alpha_{self-consistency}$ in the set $\{0, 0.01, 1 \}$ (Table~\ref{tab:regularization-ablation}). 
We can see that setting $\alpha_{self-consistency}=\alpha_{edge}=0$ is inferior to the other options.
As we would like a single set of parameters that works well, we see that selecting $\alpha_{self-consistency}=\alpha_{edge}=0.01$ appears to provide good results across the board.

\begin{table}[ht]
\begin{tabular}{c|ccc|ccc|ccc}
 $\alpha_{self-consistency}$ & \multicolumn{3}{c|}{=0} & \multicolumn{3}{c|}{=0.01} & \multicolumn{3}{c}{=1} \\
 $\alpha_{edge}$ & =0 & =0.01 & =1 & =0 & =0.01 & =1 & =0 & =0.01 & =1 \\ \hline
\begin{tabular}[c]{@{}c@{}}discrete 9 rooms\\ (GCSL: 0.6067)\end{tabular} & \begin{tabular}[c]{@{}c@{}}0.794 \\ $\pm$ \\ {[}0.005{]}\end{tabular} & \begin{tabular}[c]{@{}c@{}}0.786 \\ $\pm$ \\ {[}0.021{]}\end{tabular} & \begin{tabular}[c]{@{}c@{}}0.792 \\ $\pm$ \\ {[}0.011{]}\end{tabular} & \begin{tabular}[c]{@{}c@{}}0.776 \\ $\pm$ \\ {[}0.015{]}\end{tabular} & \begin{tabular}[c]{@{}c@{}}0.792 \\ $\pm$ \\ {[}0.011{]}\end{tabular} & \begin{tabular}[c]{@{}c@{}}0.799 \\ $\pm$ \\ {[}0.005{]}\end{tabular} & \begin{tabular}[c]{@{}c@{}}0.79 \\ $\pm$\\ {[}0.004{]}\end{tabular} & \begin{tabular}[c]{@{}c@{}}0.79 \\ $\pm$ \\ {[}0.012{]}\end{tabular} & \begin{tabular}[c]{@{}c@{}}0.796 \\ $\pm$ \\ {[}0.011{]}\end{tabular} \\
\begin{tabular}[c]{@{}c@{}}cont 4 rooms\\ (GCSL:\\ 0.6713)\end{tabular} & \begin{tabular}[c]{@{}c@{}}0.92 \\ $\pm$ \\ {[}0.007{]}\end{tabular} & \begin{tabular}[c]{@{}c@{}}0.926 \\ $\pm$ \\ {[}0.008{]}\end{tabular} & \begin{tabular}[c]{@{}c@{}}0.933 \\ $\pm$ \\ {[}0.005{]}\end{tabular} & \begin{tabular}[c]{@{}c@{}}0.918 \\ $\pm$ \\ {[}0.009{]}\end{tabular} & \begin{tabular}[c]{@{}c@{}}0.925 \\ $\pm$\\ {[}0.002{]}\end{tabular} & \begin{tabular}[c]{@{}c@{}}0.93 \\ $\pm$\\ {[}0.006{]}\end{tabular} & \begin{tabular}[c]{@{}c@{}}0.927 \\ $\pm$\\ {[}0.007{]}\end{tabular} & \begin{tabular}[c]{@{}c@{}}0.902\\ $\pm$ \\ {[}0.051{]}\end{tabular} & \begin{tabular}[c]{@{}c@{}}0.926 \\ $\pm$ \\ {[}0.004{]}\end{tabular} \\
\begin{tabular}[c]{@{}c@{}}lunar lander\\ (GCSL: \\ 0.5429)\end{tabular} & \begin{tabular}[c]{@{}c@{}}0.747 \\ $\pm$ \\ {[}0.014{]}\end{tabular} & \begin{tabular}[c]{@{}c@{}}0.748 \\ $\pm$\\ {[}0.016{]}\end{tabular} & \begin{tabular}[c]{@{}c@{}}0.746 \\ $\pm$ \\ {[}0.015{]}\end{tabular} & \begin{tabular}[c]{@{}c@{}}0.749 \\ $\pm$ \\ {[}0.009{]}\end{tabular} & \begin{tabular}[c]{@{}c@{}}0.768 \\ $\pm$ \\ {[}0.010{]}\end{tabular} & \begin{tabular}[c]{@{}c@{}}0.752 \\ $\pm$ \\ {[}0.007{]}\end{tabular} & \begin{tabular}[c]{@{}c@{}}0.744 \\ $\pm$ \\ {[}0.016{]}\end{tabular} & \begin{tabular}[c]{@{}c@{}}0.753 \\ $\pm$\\ {[}0.019{]}\end{tabular} & \begin{tabular}[c]{@{}c@{}}0.751 \\ $\pm$ \\ {[}0.008{]}\end{tabular}
\end{tabular}
\caption{
Ablation of $\alpha_{edge}$ and $\alpha_{self-consistency}$ in TraIL in various scenarios. 
GCSL success rate is listed under the scenario name for comparison.
}
\label{tab:regularization-ablation}
\end{table}

\section{Data collection with TraIL: the \textit{continuous 4 rooms} with limited start states}\label{sec:cont-4-rooms-limited}
In many GC-RL MDPs it is often the case that many states are not in the initial state distribution support.
We study this case by limiting the agent to start only in the top-left room in \textit{continuous 4 rooms - moderate noise}. 
Using the same parameters as before, we find that after training for 10k episodes, and although the models converged, the success rates for GCSL and TraIL are 0.258 $\pm$ [0.011] and 0.272 $\pm$ [0.048] respectively.
We observe that the results are much inferior to the previous experiment where the starting state was not limited, we thus allow the TraIL policy $\pi_S$ to control the data collection, and observe that after 10k more episodes, the gap to original scenario narrows (GCSL 0.483 $\pm$ [0.073], TraIL 0.496 $\pm$ [0.055]).
To assert that the performance is not attributed to training longer, we compared against a baseline agent that doesn't switch to collect data using TraIL (training 40K episodes on GCSL collected data).
We find that this model performance remains similar (and even drops by a small margin) and attains success rates of 0.227 $\pm$ [0.019] with GCSL and 0.245 $\pm$ [0.016] with TraIL.
To conclude, we found that for this scenario where the initial state distribution does not cover the entire state-space, collecting more data with TraIL provides data of greater quality allowing both GCSL and TraIL to obtain better results.

\newpage
\section{TraIL sub-procedures}

We explicitly describe the BestMode procedure (see Algorithm~\ref{alg:best-mode}).

\begin{algorithm}[ht]\caption{BestMode algorithm}
  \SetAlgoLined\DontPrintSemicolon
  \SetKwProg{myalg}{BestMode($s, g$)}{}{}
  \label{alg:best-mode}
  \myalg{ }{
  \setcounter{AlgoLine}{0}
  \nl \textbf{Inputs:} current and goal states $s, g \in S$ \\
  \nl \textbf{Output:} mode $\overline{k(s,g)}\in [K]$\\
  \nl \# $t=0$ predicts close to $s$:\\
  \nl Get GMM means for $s$: $\mu_1^s \dots \mu_k^s$ from $\pi_S(s,g,0)$ \\
  \nl \# $t=1$ predicts close to $g$:\\
  \nl Get GMM means for $g$: $\mu_1^g \dots \mu_k^g$ from $\pi_S(s,g,1)$ \\
  \nl return $\arg\min_{k\in[K]}{\left( \|\mu_k^s - s\|^2 + \|\mu_k^g - g\|^2\right)}$\\
  }
\end{algorithm}

\section{Technical Details}\label{appdx:technical-details}

In this section we provide essential technical details that was not covered in the main text.

\textbf{Code: } 
Our code was developed using PyTorch~\cite{DBLP:journals/corr/abs-1912-01703}, and the RL scenarios use the OpenAI gym\cite{brockman2016openai} package. 
The minigrid scenarios (\textit{large rooms}, \textit{double spiral}, and \textit{discrete 9 rooms}) are based on the gym-minigrid package\cite{gym_minigrid}.
Lunar Lander is based on OpenAI gym's Lunar lander\cite{brockman2016openai}.
The \textit{panda motion planning} scenario is simulated in PyBullet~\cite{coumans2019} extending the PandaGym~\cite{DBLP:journals/corr/abs-2106-13687} package.

\subsection{Behavioral Cloning Environments}

Both environments use the gym-minigrid package\cite{gym_minigrid}. In the \textit{large rooms} scenario, doors between rooms are placed in a random wall segment and are the size of a single grid cell. 
In the \textit{double spiral} the two corridors are symmetric except for a corridor on the right that break symmetry and connects the two isolated spiraling corridors.

\subsection{Behavioral Cloning Experiments}

\textbf{Data:} For both environments we collected 400 training episodes and 300 test episodes from a shortest-path planner.
We verified that no test query was present in the training set.

\textbf{Training:} we conducted 4 isolated experiments per configuration and reported their means and standard deviations. 
We initialized a replay buffer with the training demonstrations and used the same code to sample targets and update the model as in GCSL and TraIL versions in the RL experiments.
The models were trained for 80K batches in \textit{double spiral} and 160K batches in \textit{large rooms} using the Adam optimizer\cite{kingma2014adam} and batch size of 256.  

\textbf{Architecture: } all networks have two layers of 400 neurons with Relu activations\cite{agarap2018deep}.
The MDN mixture count for TraIL is 2 and the regularization coefficients are specified in the main text.

\textbf{Metrics: } 
To estimate the number of batches for each scenario we waited until the loss function and accuracy converged.
As the main text specifies, the label to measure accuracy is the first action in the reference shortest-path trajectory.

\subsection{Reinforcement Learning Environments}

We provide specific implementation details that were not covered in the main text.
\begin{enumerate}
    \item To allow accurate test time comparisons we froze a test set of start-goal pairs (which are published as assets within our code repository), and extended the gym interface to include the ability to reset from a specific start-goal pair.
    
    \item The observations of the agent are normalized to be in $[-1,1]$ in every data dimension.
\end{enumerate}

\textit{Continuous 4 rooms:}
The agent is limited to an action of norm 0.1.
A state is considered close to the goal if the difference is of norm 0.2 or less.
Noise in \textit{moderate noise} and \textit{heavy noise} is a 2D diagonal Gaussian with standard-deviation of 0.1 and 0.5 respectively. 

\textit{Lunar lander:}
Following the GCSL definition, we measure goal success by the closeness to goal in $(x,y)$ coordinate positions only. 
A state is considered close to the goal if the norm of the difference is at most 0.1 (in the original lunar lander coordinates, not the normalized state the agent sees).

\textit{Panda motion planning: }
The action in this environment is translated to a relative position to the current joints of the robot (with a limited distance of 0.03 in the normalized state-space). 
Then, the PyBullet built in position-controller takes a single simulation step using the relative position as target, and under a maximum velocity of 1.
A state is considered close to the goal if the distance in the normalized state space of the \textbf{joint positions} it is at most 0.1 (namely, like in \textit{lunar lander} we discard the velocities for measuring success).

\subsection{Reinforcement Learning Experiments}

\textbf{Training:}
Like the original GCSL we use a batch size of 256, learning rate of 5e-4, and the Adam optimizer~\cite{kingma2014adam} (same parameters for GCSL and TraIL). 
Unlike GCSL we found that for both GCSL and TraIL a replay buffer of 2K episodes works best (instead of a non-limited buffer as specified in GCSL). 
Also, we clip the TraIL gradients in the \textit{panda motion planning} to 10k.

\textbf{Architecture:}
all networks have two layers of 400 neurons with Relu activations\cite{agarap2018deep}.
The MDN mixture count for TraIL is 2 and the regularization coefficients are as specified in the main text.

\textbf{Implementation notes: }
\begin{enumerate}
    \item We also use MDN for $\pi$ when the $A$ is continuous (the original GCSL quantized continuous spaces, which is incompatible when $A$ is high dimensional and hard to find the correct resolution in other scenarios).
    
    \item As mentioned earlier, we use the suggested trajectory post-processing method to remove subsequent identical states in the data.

    \item Unlike other environments where the same test set of start-goal queries $(s,g)$ is used, in \textit{lunar lander} due to technical difficulties we were unable to fix a single set of queries and instead each evaluation randomly starts for a new query.
\end{enumerate}


\newpage






\end{document}